\documentstyle[11pt,a4wide,lsalike]{article}

\newenvironment{rei}{\begin{quotation}}{\end{quotation}}
\newcounter{sentence}
\newcommand{\sent}[1]{\refstepcounter{sentence}%
\label{#1}%
\ref{#1}}
\newcommand{\border}{\rule{\textwidth}{0.3mm}}
\newcommand{\ul}{\underline}

\title{Noun Phrase Reference\\ in Japanese-to-English Machine Translation}

\author{ Francis {\sc Bond}, Kentaro {\sc Ogura} and Tsukasa {\sc
    Kawaoka}\thanks{Now at {\bf D\=oshisha University}, Kyoto, {\sc
      Japan}: {\tt <kawaoka@wise.doshisha.ac.jp>}.} \\ \normalsize {\bf NTT Communication
    Science Laboratories} \\ \normalsize 1-2356 Take, Yokosuka-shi,
  Kanagawa-ken, {\sc Japan} 238-03 \\ \normalsize {\tt
    bond@nttkb.ntt.jp} }
\date{TMI~'95: July 1995\footnotemark[2]}

\begin{document}
\bibliographystyle{lsalike}
\maketitle

\begin{abstract}
  This paper shows the necessity of distinguishing different
  referential uses of noun phrases in machine translation.  We argue
  that differentiating between the generic, referential and ascriptive
  uses of noun phrases is the minimum necessary to generate articles
  and number correctly when translating from Japanese to English.
  Heuristics for determining these differences are proposed for a
  Japanese-to-English machine translation system.  Finally the results
  of using the proposed heuristics are shown to have raised the
  percentage of noun phrases generated with correct use of articles
  and number in the Japanese-to-English machine translation system
  {\bf ALT-J/E} from 65\% to 77\%.
\end{abstract}

\renewcommand{\thefootnote}{\fnsymbol{footnote}} \footnotetext[2]{This
  paper was presented at the Sixth International Conference on
  Theoretical and Methodological Issues in Machine Translation
  (TMI~'95) and appears in the proceedings: pp 1--14.}
\renewcommand{\thefootnote}{\arabic{footnote}}

\section{Introduction}

Determining the referential property of noun phrases is essential not
only to understanding a text, but also to decide how to generate it in
English.  This paper proposes a heuristic algorithm to determine the
referential properties of noun phrases in a Japanese text.  The
original motivation of the research was to improve the quality of
English output by NTT Communication Science Laboratories' Japanese to
English machine translation system {\bf ALT-J/E}
\cite{Ikehara:1991,Ogura:1993}.  We expect, however, that the results
will also be useful for text extraction and general text
understanding.

In this paper we use the term {\bf noun phrase reference} to describe
the relation between a noun phrase and what it stands for when it is
used.  We distinguish between three uses of noun phrases, two
referential and one non-referential.  A noun phrase can be used to
refer in two different ways: {\sc generic} where a noun phrase is used
to refer to a whole class, and {\sc referential} where a noun phrase
refers to a particular entity or entities.  A third use is {\sc
  ascriptive} where a noun phrase is used not to refer to anything but
rather, normally with a copular verb, to ascribe a property to some
referent.  Although {\sc ascriptive} noun phrases are non-referring,
we will refer to all three uses under the general term of noun phrase
reference.  This three-way distinction of noun phrase reference was
introduced in \namecite{Bond:1994} and used as a base to determine the
countability and number of noun phrases in Japanese-to-English machine
translation.  In this paper we define exactly what is meant by the
three kinds of reference and show how the distinction is essential in
the generation of articles.

This paper is structured as follows.  First, we define the three kinds
of referentiality which we distinguish and justify the definitions on
theoretical and practical grounds, comparing them with those suggested
by other researchers.  We then describe in detail a heuristic method
for determining noun phrase reference in Japanese sentences.  Next, we
show how the distinction is used in a Japanese to English machine
translation system to generate articles and number.  Finally, we look
at experimental results gained by implementing the proposed methods
and compare them to those achieved by an earlier version of the same
system, and by other systems.

\section{Definition of noun phrase reference}
\label{sec:def}

Noun phrase reference is of fundamental importance in any discussion
of meaning \cite{Lyons:1977b}.  In English, it is also important in
determining how articles should be used.  In this section we give a
more detailed definition of the three kinds of noun phrase reference
under discussion and compare them with the definitions used in other
machine translation systems.

\begin{description}
\item[Generic:] Noun phrases with generic reference denote an entire
  class: e.g. {\it mammoths\/} in {\it \ul{Mammoths} are extinct}.  In English 
  generic noun phrases can normally be expressed in three ways, as
  discussed in Section~\ref{sec:gen}.

\item[Referential:] Referential noun phrases are those that refer to
  some entity or entities in the discourse world: e.g. {\it a mammoth\/}
  in {\it There is \ul{a mammoth} in my garden!\/}  Referential noun phrases 
  are plural if there is more than one discrete referent, and are
  marked for definiteness.  
\item[Ascriptive:] Ascriptive noun phrases are
  used with a copular verb, or in an appositive expression, to ascribe
  a property to their subject: e.g. {\it a mammoth\/} in {\it That animal
    is \ul{a mammoth}}.  Because ascriptive noun phrases are non-referring
  they cannot be the antecedent of other noun phrases.
\end{description}


\namecite{Zel:1992} distinguishes between {\sc generic} and {\sc
  identifying}, which appear to be equivalent to our {\sc generic} and
{\sc referential}.  \citeauthor{Zel:1992}'s examples do contain
ascriptive noun phrases, for example {\it a human being\/} in {\it `A
  spectator is a human being'}, instead they appear to be treated as
adjective phrases in the rules (for example in their rule 14 (p. 797
op cit) where the complement of the copulative predicate with a
generic subject is an evaluative adjective phrase).  If the definition
of adjective phrase has been expanded to include {\sc ascriptive} noun
phrases\footnote{We feel this expanded definition is plausible, since
  the copula and ascriptive noun phrase combination fulfills the same
  semantic role as the copula and adjective phrase, that is, to
  ascribe a property.} then our analysis is compatible.  Unfortunately
there is no discussion in \citeauthor{Zel:1992} as to how effective
their rules are when actually used in a machine translation system so
we cannot make a quantitative comparison.

\namecite{Murata:1993b} distinguish between {\sc generic} and {\sc
  non-generic}, which is further divided into {\sc definite} and {\sc
  indefinite}, using heuristics similar to rewriting rules in expert
systems.  They make no distinction between {\sc referential} and {\sc
  ascriptive} for non-generic noun phrases.  This leaves open the
possibility for conflict with their rule that a noun phrase will be
definite if it has been presented previously.  Consider the following
sentence\footnote{Examples are given with the (romanized) Japanese
  original, a gloss and the human translation.  The examples have been
  simplified to exemplify points more clearly; a new translation has
  been made for each simplified sentence.  Japanese particles are
  glossed as follows: {\sc top} for {\it wa\/} which marks the topic,
  {\sc obj} for {\it o\/} which marks the object and {\sc gen} for
  {\it no\/} which shows a genitive relation.}: {\it z\=o-wa
  hony\=urui da-si, manmosu-mo hony\=urui da.\/} `Elephant-{\sc top}
mammal be-and mammoth-{\sc also} mammal be.' {\it Elephants are
  mammals and mammoths are also mammals}.  This will become {\it
  Elephants are mammals and mammoths are also \ul{the} mammals\/}
using the rules given.  Distinguishing between {\sc referential} and
{\sc ascriptive} prevents this kind of problem from occurring.  We
compare their results to ours quantitatively in Section~\ref{sec:exp}.

\section{Determination of noun phrase reference}
\label{sec:det}

All proper nouns are, by definition, {\sc referential}.  The algorithm
used to determine the referential property of noun phrases headed by
common nouns is shown in Figure~\ref{fig:ref}.  The algorithm
presented is based on single sentences, it does not address the
considerable problems of using information from outside the sentence
being considered\footnote{Algorithms to use contextual information
  from outside the sentence are currently being implemented.}.


It is possible for the algorithm to be applied to the Japanese parse
tree as part of the semantic analysis\footnote{For information
  retrieval it is obviously essential to determine the referentiality
  of noun phrases as part of the source language analysis.}.  In {\bf
  ALT-J/E}, however, the algorithm is applied after the semantic
analysis has finished, during the transfer stage, because much of the
semantic information is stored in the transfer dictionaries where the
combination of Japanese and English makes it easy to disambiguate word
senses.  The overall process of translation in {\bf ALT-J/E} is
divided into seven parts.  First, the system splits the Japanese text
into morphemes and assigns parts of speech.  Second, it parses the
segmented text, often giving multiple possible interpretations.
Third, it rewrites complicated Japanese expressions into simpler ones.
Fourth, {\bf ALT-J/E} semantically evaluates the various
interpretations.  Fifth, syntactic and semantic criteria are used to
select the best interpretation.  Sixth, the selected interpretation is
transferred into English.  Finally, the English sentence is adjusted
to give the correct inflectional forms. The algorithm described in
this section has been implemented as part of the sixth stage.
However, it could be implemented as part of the fifth stage.

Rules are applied in the order shown in Figure~\ref{fig:ref}, with
later rules over-ruling earlier ones.


\begin{figure}[htpb]
\border
\begin{center}
\small
\begin{enumerate}
\item The default is {\sc referential}
\item Sentence level rules \label{lev:sent}
  \begin{enumerate}
  \item the subject of a verb marked in the lexicon as predicating over an entire class is
    {\sc generic}:\\ {\it \ul{manmosu-wa} zetsumetsu-shita\/}
    `\ul{Mammoths} died out'
  \item \label{det:z} if the semantic category of the subject of a
    copula is a descendant of the semantic category of the object then
    the subject is {\sc generic}:\\ {\it  \ul{manmosu-wa}
      d\=obutsu-da\/} `\ul{Mammoths} are animals'
  \item the object of a verb which predicates {\sc emotive action} or
  {\sc emotive state} is {\sc generic}: \\
  {\it watashi-wa \ul{manmosu-wo} suki-da\/} `I like \ul{mammoths}'
  \item the complement of a copula is {\sc ascriptive}:
    \\ {\it manmosu-wa \ul{d\=obutsu-da\/}} `Mammoths are \ul{animals}'
  \item appositive noun phrases are {\sc ascriptive}:\\ 
  {\it \ul{denwagaisha-no} NTT\/} `NTT, \ul{a telephone company}'
  \end{enumerate}
\item Modification by embedded sentences \label{lev:embed}
  \begin{enumerate}
  \item A noun phrase whose head is modified by a tensed relative
    clause is {\sc referential}: \\ 
    {\it \ul{kinou kita otoko}\/} `\ul{the man who came yesterday}'
  \end{enumerate}
\item Post-modification by {\it setsubiji\/} `suffixes' and {\it
    joshi-s\=ot\=ogo\/} `pseudo-particles' \label{lev:post}
  \begin{enumerate}
  \item the modificant of {\it muke\/} `aimed at', {\it y\=o\/} `for'
      \ldots is {\sc generic}:\\ 
      {\it \ul{josei-muke}-no zasshi\/} `A magazine \ul{for women}'
  \item the modificant of {\it -to-iu-no-wa \/} `things called' is
    {\sc generic}: \\
    {\it \ul{kikai hon'yaku-to-iu-no-wa} muzukashii\/} 
    `\ul{Machine translation} is difficult'
  \end{enumerate}
\item Modification by demonstratives, numerals and the genitive construction
  {\it no\/} `of' \label{lev:this}
  \begin{enumerate}
  \item A noun phrase whose head is  modified by a demonstrative or
    numeral is {\sc referential}: \\
    \ul{\it kono otoko\/} `\ul{this man}', \ul{\it futari-no otoko\/} `\ul{two men}'
  \item A noun phrase whose head is modified by the genitive construction is
    {\sc referential}: \\   
    \ul{\it hana-no saki\/} `\ul{the tip of my nose}'
  \end{enumerate}
  \item A noun phrase with a `unique' referent is {\sc referential}:
    \label{lev:uniq} \\ 
    \ul{\it chiky\=u\/} `\ul{the earth}'
\end{enumerate} 
\end{center}
\caption{Determination of noun phrase referentiality}
\label{fig:ref} 
\border
\end{figure}

The default assumption is that a noun phrase will be used to refer to
some specific entity or entities in the discourse world, i.e. that it is 
{\sc referential}.  

There are five rules that are applied at the sentence level, which use
the meanings of verbs combined with the semantic categories of
nouns\footnote{The meanings of nouns are given in terms of a semantic
  hierarchy of 2,800 nodes.  Each node is called a semantic category.
  Edges in the hierarchy represent {\sc is-a} relationships, so that
  the child of a semantic category {\sc is-a} instance of it.  For
  example, {\sc organ is-a body-part} \cite{Ogura:1993}.}.  These
can all be overridden by subsequent rules.  The subjects of verbs that
predicate over an entire class, and the objects of verbs which
predicate {\sc emotive action} or {\sc emotive state}, are {\sc
  generic}.  Verbs that trigger these rules, e.g. {\it evolve, die
  out\/} are marked in the lexicon \cite{Bond:1993b}.  For copulas,
the subject is {\sc generic} if its semantic category is a descendent
of the semantic category of the object, while it's complement is taken
to be {\sc ascriptive} by default\footnote{If the complement is later
  judged to be {\sc referential} by a subsequent rule it is equivalent
  to judging that the copula has been used equatively.}.  Finally,
appositive noun phrases will be judged to be {\sc ascriptive}, as
though they were the complement of a copula.

Recall that these rules are only applied if the noun phrase in
question is headed by a common noun.  In sentence~\ref{sent:aoi-hall},
the semantic category of {\it meeting place\/} is {\sc actual place},
which is a child of the semantic category of {\it Aoi hall} {\sc
  public place}. 
{\it Aoi hall}, however, is a proper noun so the rule is not applied.

\begin{rei} 
\begin{tabular}{cllll}
(\sent{sent:aoi-hall})
    & Jap:   &  \it kaij\=o-wa &  \it Aoi-kaikan & $\phi$.   \\
    & Gloss: &  \tt meeting place-{\sc top} & \tt Aoi hall &is  \\
    & Eng:   & \multicolumn{3}{l}{The meeting place is the Aoi Hall} \\
\end{tabular}
\end{rei}

The next level of rules (level~\ref{lev:embed}) applies to noun
phrases modified by embedded sentences.  Japanese makes no
phonological, morphological, or syntactic distinctions between
restrictive and non-restrictive relative clauses \cite[235]{Kuno:1973}.
This algorithm uses a simple heuristic: a noun phrase modified by a
tensed embedded sentence is {\sc referential}.

The next level of rules (level~\ref{lev:post}) is based on
post-modification in the Japanese sentence.  The use of some {\it
  setsubiji} `suffixes'\footnote{{\it setsubiji\/} are a Japanese part
  of speech made up of suffixes that cannot stand alone, but change
  the meaning of the word they modify.} implies that their modificant
is {\sc generic}.  For example {\it muke\/} `aimed at' in {\it
  josei-muke-no-zasshi\/} `woman aimed-at {\sc gen} magazine' {\it a
  magazine aimed at women}.  Similarly the construction {\it
  A-to-iu-no-wa\/} `things called A' implies that its modificant is
{\sc generic}.  It can in fact be thought of as a pseudo-particle, the
whole construction acting as a single marker which has the effect of
marking it's modificant as being a generic noun phrase used as the
topic\footnote{In {\bf ALT-J/E} the entire construction (and the
  similar construction {\it A-to-iu-mono-wa\/} `things called A') is
  rewritten during the Japanese rewriting stage into a pseudo-particle
  \cite{Shirai:1993}, which marks its modificant as being a generic
  noun phrase in the {\it ha}-case ({\sc topic}).  It is not however
  necessary to do this, as shown in \namecite{Murata:1993a}, where this
  construction is found by matching against the Japanese dependency
  structure.}.

The next level of rules (level~\ref{lev:this}) makes a noun phrase
whose head is modified by a demonstrative, numeral or the genitive
construction {\it NP-no\/} `NP's' {\sc referential}.  Note that only
noun phrases modified by {\it no\/} judged to be genitive are {\sc
  referential}.  Partitive constructions such as {\it
  \=okami-no-mure\/} `pack of wolf' {\it a pack of wolves\/} are not
included in this judgment.  The genitive construction may be
translated into English in a variety of ways including a prepositional
phrase headed by `of', a possessive phrase with a clitic in the
determiner position, or a possessive pronoun. 

Finally (level~\ref{lev:uniq}), noun phrases headed by nouns that are
marked in the lexicon as likely to have a unique referent, such as
{\it chiky\=u\/} `the earth' are assumed to be {\sc referential}.

The algorithm presented in this section is only heuristic.  Further
work remains to be done to refine it.  In particular: using the wa/ga
distinction in conjunction with noun anaphora relations to distinguish 
between {\sc generic} and {\sc refer\-ential}, and improving the rules at
level~\ref{lev:embed} for relative clauses.

\section{Using noun phrase referentiality to select articles and
  determine number}

Knowledge of a noun phrase's referential use is essential when
translating from Japanese to English, as it plays a large part in
determining how a noun phrase is expressed in English.  In this
section we show how articles and number are generated differently for
the three different referentialities in the machine translation system
{\bf ALT-J/E}.  Correct generation of articles and number is important
not only to express meaning accurately, but because it is one of the
major factors in determining the readability of Japanese-to-English
translations.

\subsection{Translation of generic noun phrases}
\label{sec:gen}

A {\sc generic} noun phrase (with a countable head noun) can generally
be expressed in three ways \cite{Huddleston:1984}.  We call these {\sc
  gen} `a', where the noun phrase is indefinite: {\it \ul{A mammoth}
  is a mammal\/}; {\sc gen} `the', where the noun phrase is definite:
{\it \ul{The mammoth} is a mammal\/}; and {\sc gen}~$\phi$, where
there is no article: {\it \ul{Mammoths} are mammals}.  Uncountable
nouns and pluralia tantum can only be expressed by {\sc gen}~$\phi$
(eg: {\it \ul{Furniture} is expensive\/}).  They cannot take {\sc gen}
`a' and they do not take {\sc gen} `the', because then the noun phrase
would normally be interpreted as having definite reference.  Nouns
that can be either countable or uncountable take only {\sc gen}~$\phi$
or `a': {\it \ul{Cake} is delicious}/{\it \ul{Cakes} are delicious},
{\it \ul{A cake} is a kind of food\/}.  These combinations are shown
in Table~\ref{gandc}.  Noun phrases that cannot be used to show {\sc
  generic} reference are marked with an asterisk (*).

\begin{table}[htp]
  \border
    \caption{Genericness and Countability}
    \label{gandc}
    \begin{center}
    \begin{tabular}{c|lll} 
      GEN & \multicolumn{3}{|c}{Noun Countability Preference} \\ 
      type & Countable & Both & Uncountable \\ \hline
      `a' & a mammoth & a cake & *a furniture \\ 
      `the' & the mammoth & *the cake & *the furniture \\ 
      $\phi$ & mammoths & cake/cakes & furniture
    \end{tabular}
      
    \end{center}
  \border
\end{table}

The use of all three kinds of {\sc generic} noun phrases is not acceptable
in some contexts, for example *{\it \ul{a mammoth} evolved}.  Sometimes a
noun phrase can be ambiguous, for example {\it I like \ul{the elephant}},
where the speaker could like a particular elephant, or all elephants.

Because the use of {\sc gen}~$\phi$ is acceptable in all contexts, {\bf
  ALT-J/E} generates all {\sc generic} noun phrases as such, that is as bare
noun phrases.  The number of the noun phrase depends on the
countability preference of the noun phrase heading it and there will
be no article.

\subsection{Translation of referential noun phrases}
\label{sec:ref}

The countability and number of {\sc referential} noun phrases can be
determined with heuristics that use information from the Japanese
sentence along with knowledge of English countability stored in the
lexicon.  This is described in \namecite{Bond:1994}.

According to \namecite[265]{Quirk:1985}, for {\sc referential} noun phrases: 
\begin{quote}
  The definite article {\it the\/} is used to mark the phrase it
  introduces as referring to something which can be identified
  uniquely in the contextual or general knowledge shared by speaker
  and hearer.
\end{quote}

Whether or not a {\sc referential} noun phrase is definite or not is
determined using heuristic criteria based on whether there is enough
information to uniquely identify the noun phrase's referent, such as
the following:

\begin{itemize}
\item if the head noun is marked in the lexicon as being unique: \\ 
  {\it the earth\/}
\item if the noun phrase is made logically unique by a modifier: \\ 
  {\it the best price\/}
\item if the noun phrase's referent is restrictively described: \\
  {\it the man who came to dinner}, {\it the aim of this research\/}
\item direct and indirect anaphoric reference: \\
  {\it I saw a cat and a dog.  \ul{The dog} chased \ul{the cat}.\/}
\end{itemize}

As the above criteria are only meaningful for {\sc referential} noun
phrases, it is essential to determine whether the noun phrase is
referential as a first step.

When it has been determined whether a noun phrase is definite or
indefinite, then articles can be generated\footnote{As well as
  generating definite and indefinite articles, {\bf ALT-J/E} also
  generates possessive pronouns \cite{Bond:1995a} and {\it some/any\/}
  for {\sc referential} noun phrases when appropriate.}.  In the final
stage of processing, if there is no determiner, definite noun phrases
take the definite article {\it the}.  Indefinite countable singular
noun phrases will take the indefinite article {\it a/an}, while
indefinite countable plural and uncountable noun phrases will take the
zero article $\phi$.  This is summarized in Table~\ref{tab:art}.

\begin{table}[htbp]
  \border
    \caption{Generation of articles for referential noun phrases.}
    \label{tab:art}
    \begin{center}
    \begin{tabular}{lcc}
      Noun Phrase Number &  Definite &  Indefinite \\ \hline
      Countable singular & \it the & \it a/an \\
      Countable plural & \it the  & $\phi$ \\
      Uncountable & \it the  & $\phi$ 
    \end{tabular}
    \end{center}
  \border
\end{table}

\subsection{Translation of ascriptive noun phrases}
\label{sec:asc}

The countability and number of {\sc ascriptive} noun phrases matches
that of their subject, and the countability and number of two
appositive noun phrases match each other as described in
\namecite{Bond:1994}, with the following proviso.  If one element is
plural and the other is a collective noun such as {\it group}, then
they need not match.  For example, {\it many insects, a whole swarm,
  \ldots\/} as opposed to {\it many insects, bees I think, \ldots\/}.

{\bf ALT-J/E} makes the simplifying assumption that all {\sc ascriptive}
noun phrases are indefinite.  Therefore, articles will be generated in
the same way as for indefinite {\sc referential} noun phrases.  Countable
singular noun phrases will therefore take the indefinite article {\it
  a/an}, and countable plural and uncountable noun phrases will take
the zero article $\phi$.

\section{Results}
\label{sec:exp}

The processing described above has been implemented in {\bf ALT-J/E}.
The rules were designed using data from a specially constructed set of
test sentences collected by the authors.  The algorithm was evaluated
on a collection of newspaper articles from the {\it Nikkei-Sangyou\/}
newspaper by an English native speaker not connected with the
development of the algorithm.  The results are summarized in
Table~\ref{tab:overall}.

\begin{table}[htbp]
  \border
  \caption{Correct Generation of Articles and Number}
  \begin{center}
  \label{tab:overall}
  \small 
  \begin{tabular}{lcccc} 
    & \multicolumn{2}{c}{Test Sentences} 
    & \multicolumn{2}{c}{Newspaper Articles} \\ 
    & NPs  (240)  & Sentences   (120)   & NPs  (717)  & Sentences (102) \\ \hline
    New:    & 94\%  & 90\%          & 77\%  & 15\%  \\ 
    Old:    & 70\%  & 46\%          & 65\%  & 5\%  \\ 
  \end{tabular}

  New shows the results using the proposed method.

  Old shows the results using the unmodified system.
\end{center}\border
\end{table}


We tested the system on newspaper articles, in the articles tested,
there were an average of 7 noun phrases in each sentence.  The
articles were translated by {\bf ALT-J/E} and the raw output examined
by an English native speaker.  Each noun phrase was given one of the
following scores:
\begin{description}
\item[{\sc structure}:] problem with structure or choice of
  translation\footnote{This includes any major problems not connected
    with articles or number, such as outputing Japanese characters or
    spelling errors.}
\item[{\sc best}:] the most appropriate article/number
\item[{\sc article}:] inappropriate article
\item[{\sc number}:] inappropriate number
\item[{\sc possessive}:] inappropriate use of possessive determiner
\item[{\sc countability}:] problem with countability
\item[{\sc reference}:] problem with referential property
\end{description}
For the purpose of evaluating the generation of articles and number,
noun phrases that were either the {\sc best} possible translation, or
that had a problem only with {\sc structure/choice of translation},
were judged to be successful.  A third-party evaluator gave the
success rates as 77\% for the system with the proposed method and 65\%
for the original system.  The method of evaluation described above
does not give a reproducible, absolute level of success.  It does,
however, successfully show the overall level of
improvement/degradation, and help to identify the remaining problems.

Our initial evaluation was done by the the authors, who found the
success rates at the noun phrase level to be 92\% for the proposed
method and 76\% for the system as it used to be.  Nakazawa points out
that this shows that the evaluation method is not reproducible
(personal communication May 1995).  Because the goal is to produce a
translation, which is new text, there is no objective target to
compare the results with.  This is a perennial problem for machine
translation output.  \namecite{Knight:1994} in a small pilot study
showed that humans could replace articles ({\it a/an\/} and {\it
  the\/}) in an English text in which the articles had been replaced
by blanks with an accuracy of around 95\%.  Raw machine translation
output is less coherent than normal English text and so deciding which
article is appropriate is an even harder task.

\section{Discussion}
\label{sec:disc}

In this section we discuss the remaining errors and compare
the results to two other systems.

168 of the 717 noun phrases in the machine translation of the
newspaper articles had some problem. An brief analysis of the errors
is given in Table~\ref{tab:errors}.

\begin{table}[htbp]
  \border
  \caption{Errors in the generation of articles and
    number}
  \label{tab:errors}
 \begin{center}
  There were 168 errors in the 717 noun phrases \\that appeared in the
  machine translation of the newspaper articles
  \end{center}
  \begin{center}
    \begin{tabular}{lcp{8cm}}
      Problem Area & Freq. & Description of error\\ \hline \hline
      Analysis error & 22\% & The Japanese noun phrase was parsed incorrectly 
      so the rules did not trigger. \\
      Dictionary errors & 22\% & The dictionary entry was incomplete.\\
      Numerical  & 12\% & Complicated numerical expressions are \\
      Expressions & & translated badly: for example \ul{\it 384 Kbits of 
        networks per second\/} should be \ul{\it a 384 Kbit/s network\/}
      \\ \hline
      Reference & 8\% & There needs to be a rule to make {\it
        database\/} {\sc generic} in expressions like: {\it the
        strategic applications of \ul{databases}} which is currently
      translated as  {\it the strategic applications of \ul{a
          database}}\\ 
      Reference & 5\% & Miscellaneous errors in determining noun
      phrase reference. \\ \hline
      Number & 9\% & In some cases rules using common sense and inference are
      needed to determine the number correctly: for example {\it sales 
        counter\/} should be plural in {\it \ul{the sales counter} of
        telephone companies through out the country\/} \\
      Number & 2\% &  There are no rules to deduce number from
      information given by adverbs: for example {\it prices\/} should
      be plural in {\it \ul{The  price} is 5 yen and 15 yen
        respectively\/} \\ \hline
      Articles & 7\% & The rules for deciding whether a noun has been
      restrictively described by an embedded sentence are too coarse.  \\
      Articles & 6\% & There needs to be a rule for indirect
      anaphora.  {\it two models\/} should be definite in {\it NTT
        introduced video-tel 111 and video-tel 222 in June.  \ul{Two models} are
        the first to have video receivers.\/} \\
      Articles & 3\% & There needs to be a rule to make a noun phrase
      definite if its pre-head modifier restricts it sufficiently: for
      example  {\it NTT will enter \ul{a} video rental business\/} \\
      Articles & 4\% & Miscellaneous errors in determining whether a
      noun phrase is definite or not. 
    \end{tabular}
  \end{center}
  \border
\end{table}

Testing on the newspaper articles revealed one major heuristic that
had been overlooked in the algorithm presented in
section~\ref{sec:det}: some nouns when heading a construction such as
`N-{\it of\/}-NP' carry an implication that the complement NP has {\sc
  generic} reference: for example, {\it the applications of
  \ul{databases}}.  This rule will be added to the algorithm at
level~\ref{lev:this}, reducing the number of errors by around 8\%.
Apart from this there were no major changes that needed to be made to
the algorithm.

Overall, the largest sources of errors are problems with the source
language analysis and dictionaries (22\% each).  These are not
problems with the proposed algorithm but with the machine translation
system as a whole.  Another major source of errors is the translation
of numerical expressions (12\%).  The processing for handling
numerical expressions is currently being overhauled.  The errors
caused by lack of information in the dictionaries are solvable
immediately, which will reduce the number of errors by around 20\%.

In the generation of articles and numbers for {\sc referential} noun
phrases some of the errors can simply be solved by the addition of new
rules: for example, adding rules which use the meaning of adverbs to
determine number or rules using pre-head modifiers to determine
definiteness.  The problems of common sense deduction and indirect
anaphora, however, require a large scale knowledge base and inference
rules.  While both are being researched at the moment, they are
unlikely to be implemented soon.  We estimate that the number of
errors caused by insufficiencies in the generation of articles and
numbers for {\sc referential} noun phrases can be reduced at least a
quarter, thus reducing the total number of errors by around 8\%.

Combining the above figures, we predict it is possible to reduce the
errors by around 30\%, bringing the total success rate to 84\% for a
window test.  To go beyond this needs new processing to improve the
source language analysis, the translation of numerical expressions and
more use of contextual inferences.

In addition examining even this small sample of text we came up with
one major addition to the algorithm for determining noun phrase
reference.  Therefore the algorithm needs to be tested on a wider
range of texts before the rules can be considered comprehensive.  We
have started testing the algorithm on a larger corpus of newspaper
articles and are investigating methods for automatically learning
rules.

In \namecite{Murata:1993b} success rates of 68.9\% for referential
property and 85.6\% for number were given for unknown texts of the
same genre as that used in development of the rules.  Their approach
seems effective, although we predict the lack of a {\sc ascriptive}
class will cause problems.  It is impossible to directly compare our
results as \citeauthor{Murata:1993b}'s testing was all carried out in
Japanese by the developers, so the problems of actually generating the
English and getting an impartial evaluation were not addressed.
Setting these considerations aside, when we separate our results for
noun phrase reference (counting as failures noun phrases with errors
in article use, noun phrase reference or the use of possessive
determiners), and countability and number (counting as failures noun
phrases with errors in number or countability), our proposed algorithm
gave success rates of 74\% and 85\% respectively.

Another approach is that of \cite{Knight:1994}, who proposed
using an automated post-editor to correct articles.  Their prototype
has a success rate for learning to replace articles when they have
been removed from English texts of 78\%.  At present however the
prototype cannot be used to post-edit output from a typical machine
translation system as it assumes the knowledge that an article should
be used in a given position, which is not normally available, and that
the generation rules can function using machine translation output,
which has not been shown.



\section{Conclusion}
\label{sec:conc}

This paper proposes a method that uses the information available in a
Japa\-nese sentence to identify a noun phrase as being used either
{\sc generically}, {\sc referentially} or {\sc ascriptively}.  This
distinction is shown to be both theoretically justified and
practically useful.  The three way distinction in noun phrase
reference is used as a base to determine a noun phrase's number and to
generate appropriate articles and possessive pronouns when translating
from Japanese to English.  Incorporating this method into the machine
translation system \mbox{{\bf ALT-J/E}} helped to improve the
percentage of noun phrases with correctly generated articles and
number from 65\% to 77\%.  It is shown that the proposed method can be
extended straightforwardly to increase the success rate to 84\%.

Several problems remain to be explored.  We consider the following to
of primary importance:
\begin{enumerate}
\item Extension of the algorithm to translate texts as coherent passages,
  not just as single sentences.
\item Improvement of the reproducibility of the evaluation method.
\item Investigation of the coverage of the algorithm on a wider
  collection of texts.
\end{enumerate}

\section*{Acknowledgments}

The paper has benefited greatly from the comments of the anonymous
reviewers for TMI, Graham, Monique and Mitsuyo Bond, Satoru Ikehara,
Roly Sussex and especially Tsuneko Nakazawa.  We would like to thank
Toshiaki Nebashi, Kazuya Fukamachi and Yoshitake Ichii for their
invaluable help in implementing the processing described here.

\end{document}